%% file: main.tex
\documentclass[10pt,twocolumn,letterpaper]{article}

\usepackage[pagenumbers]{cvpr} 

\input{preamble}

\definecolor{cvprblue}{rgb}{0.21,0.49,0.74}
\usepackage[pagebackref=false,breaklinks=true,colorlinks,bookmarks=false]{hyperref}
\hypersetup{linkcolor=[rgb]{0.7,0.1,0.1}}
\hypersetup{citecolor=[rgb]{0.4,0.15,0.95}}
\hypersetup{urlcolor  = cvprblue}

\usepackage{xcolor,colortbl}
\definecolor{Gray}{gray}{0.94}

\usepackage{tocloft}
\usepackage[toc,page,header]{appendix}
\usepackage{adjustbox}

\usepackage{minitoc}

\renewcommand \thepart{}
\renewcommand \partname{}

\newlength\savewidth\newcommand\shline{\noalign{\global\savewidth\arrayrulewidth
\global\arrayrulewidth 1pt}\hline\noalign{\global\arrayrulewidth\savewidth}}

\input{shortcuts}

\def\paperID{7330} 
\def\confName{CVPR}
\def\confYear{2024}

\title{\vspace{-3mm}GraphDreamer: Compositional 3D Scene Synthesis from Scene Graphs}

\author{\fontsize{11.25pt}{\baselineskip}\selectfont Gege Gao\textsuperscript{1,2,3,4}~~~Weiyang Liu\textsuperscript{1,5,*}~~~Anpei Chen\textsuperscript{2,3,4}~~~Andreas Geiger\textsuperscript{3,4,\textdagger}~~~Bernhard Schölkopf\textsuperscript{1,2,4,\textdagger}\\[0.75mm]
\fontsize{11.25pt}{\baselineskip}\selectfont \textsuperscript{1}Max Planck Institute for Intelligent Systems -- Tübingen~~~~\textsuperscript{2}ETH Zürich~~~~\textsuperscript{3}University of Tübingen\\[0.5mm]
\fontsize{11.25pt}{\baselineskip}\selectfont\textsuperscript{4}Tübingen AI Center~~~~\textsuperscript{5}University of Cambridge~~~~\textsuperscript{*}Directional lead~~~~\textsuperscript{\textdagger}Shared last author
}

\begin{document}

\doparttoc 
\faketableofcontents

\twocolumn[{

    \renewcommand\twocolumn[1][]{#1}
    \maketitle

\vspace{-9.5mm}
\begin{center}
        \fontsize{10.25pt}{\baselineskip}\selectfont
        {\tt\href{https://graphdreamer.github.io}{\textbf{graphdreamer.github.io}}}
\end{center}
\vspace{.4mm}

    \begin{center}
        \input{figures/teaser}
    \end{center}
}]

\input{sec/0_abstract}    
\input{sec/1_intro}

\input{sec/2_related}
\input{sec/3_preliminary}

\input{sec/4_method}
\input{sec/5_experiments}
\input{sec/6_conclusion}
\input{sec/7_acknowledgments}
{
    \small
    \bibliographystyle{ieeenat_fullname}
    \bibliography{bibliography_long,bibliography,bibliography_custom}
}

\input{sec/X_suppl}

\end{document}

%% file: preamble.tex
\usepackage{caption}

\usepackage[dvipsnames]{xcolor}

\usepackage{graphicx}
\usepackage{amsmath}
\usepackage{amssymb}
\usepackage{booktabs}
\usepackage{url}
\usepackage{epsfig}
\usepackage{lipsum}
\usepackage{amsthm}
\usepackage{comment}
\usepackage{multirow}
\usepackage{subcaption}
\usepackage{microtype}
\usepackage{xspace}
\usepackage{setspace}
\usepackage[dvipsnames]{xcolor}
\usepackage[utf8]{inputenc}
\usepackage{enumitem}
\usepackage{makecell}
\usepackage{float}
\usepackage{bm}
\usepackage{commath}
\usepackage{wrapfig,lipsum,booktabs}

\definecolor{myred1}{RGB}{191, 82, 63}
\definecolor{myblue1}{RGB}{84, 138, 200}
\definecolor{mygreen1}{RGB}{70, 162, 54}

\makeatletter
  \newcommand\figcaption{\def\@captype{figure}\caption}
  \newcommand\tabcaption{\def\@captype{table}\caption}
\makeatother

%% file: shortcuts.tex
\newcommand{\bd}{\mathbf{d}}

\newcommand{\bo}{\mathbf{o}}
\newcommand{\bp}{\mathbf{p}}

\newcommand{\br}{\mathbf{r}}
\newcommand{\bs}{\mathbf{s}}

\newcommand{\bu}{\mathbf{u}}

\DeclareMathOperator*{\argmax}{argmax~}

\makeatletter
\DeclareRobustCommand\onedot{\futurelet\@let@token\@onedot}
\def\@onedot{\ifx\@let@token.\else.\null\fi\xspace}
\def\eg{\emph{e.g\onedot}} 
\def\ie{\emph{i.e\onedot}}

\makeatother

\definecolor{darkgreen}{rgb}{0,0.7,0}


%% file: figures/teaser.tex
    \vspace{-2mm}
    \centering
    \includegraphics[width=1\linewidth]{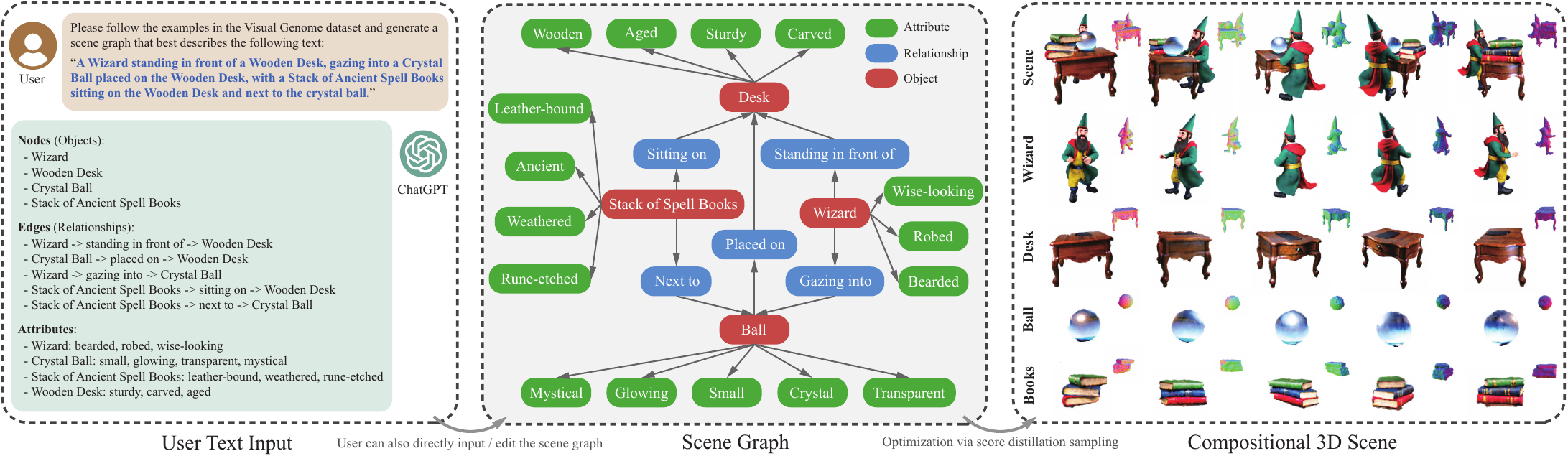}
    \vspace{-4.5mm}
    \captionof{figure}{
        \footnotesize \emph{GraphDreamer} takes a scene graph as input and generates a compositional 3D scene where each object is fully disentangled. To save the effort of building a scene graph from scratch, the scene graph can be generated by a language model (e.g., ChatGPT) from a user text input (left box).
    }
    \label{fig:teaser}
    \vspace{0.6cm}

%% file: sec/0_abstract.tex
\begin{abstract}
\vspace{-3mm}
As pretrained text-to-image diffusion models become increasingly powerful, recent efforts have been made to distill knowledge from these text-to-image pretrained models for optimizing a text-guided 3D model. Most of the existing methods generate a holistic 3D model from a plain text input. This can be problematic when the text describes a complex scene with multiple objects, because the vectorized text embeddings are inherently unable to capture a complex description with multiple entities and relationships. Holistic 3D modeling of the entire scene further prevents accurate grounding of text entities and concepts. To address this limitation, we propose GraphDreamer, a novel framework to generate compositional 3D scenes from scene graphs, where objects are represented as nodes and their interactions as edges. By exploiting node and edge information in scene graphs, our method makes better use of the pretrained text-to-image diffusion model and is able to fully disentangle different objects without image-level supervision. To facilitate modeling of object-wise relationships, we use signed distance fields as representation and impose a constraint to avoid inter-penetration of objects. To avoid manual scene graph creation, we design a text prompt for ChatGPT to generate scene graphs based on text inputs. We conduct both qualitative and quantitative experiments to validate the effectiveness of GraphDreamer in generating high-fidelity compositional 3D scenes with disentangled object entities.
\end{abstract}

%% file: sec/1_intro.tex
\vspace{-7mm}
\section{Introduction}
\vspace{-.2mm}

Recent years have witnessed substantial progresses in text-to-3D generation~\cite{poole2022dreamfusion, wang2022score, metzer2022latentnerf}, largely due to the rapid development made in text-to-image models~\cite{rombach2021highresolution,saharia2022photorealistic} and text-image embeddings~\cite{radford2021learning}. This emerging field has attracted considerable attention due to its significant potential to revolutionize the way artists and designers work.

The central idea of current text-to-3D pipelines is to leverage knowledge of a large pretrained text-to-image generative model to optimize each randomly sampled 2D view of a 3D object such that these views resemble what the input text describes. The 3D consistency of these 2D views is typically guaranteed by a proper 3D representation (\eg, neural radiance fields (NeRF)~\cite{Mildenhall2020ECCV} in DreamFusion~\cite{poole2022dreamfusion}). Despite being popular, current text-to-3D pipelines still suffer from \emph{attribute confusion} and \emph{guidance collapse}. Attribute confusion is a fundamental problem caused by text-image embeddings (\eg, CLIP~\cite{radford2021learning}). 
For example, models often fail at distinguishing the difference between ``a black cat on a pink carpet'' and ``a pink cat on a black carpet''. This problem may prevent current text-to-3D generation methods from accurately grounding all attributes to corresponding objects. As the text prompt becomes even more complex, involving multiple objects, attribute confusion becomes more significant. Guidance collapse refers to the cases where the text prompt is (partially) ignored or misinterpreted by the model. This typically also happens as the text prompt gets more complex. 
For example, ``a teddy bear pushing a shopping cart and holding baloons'', with ``teddy bear'' being ignored.
These problems largely limit the practical utility of text-to-3D generation techniques. 

A straightforward solution is to model the multi-object 3D scene in a compositional way. 
Following this insight, recent methods \cite{lin2023componerf,po2023compositional,Dhamo2021graphto3D,zhai2023commonscenes} 
condition on additional context information such as 3D layout which provides the size and location of each object in the form of non-overlapping 3D bounding boxes. 
While a non-overlapping 3D layout can certainly help to produce a compositional 3D scene with each object present, it injects a strong prior and greatly limits the diversity of generated scenes. The non-overlapping 3D box assumption can easily break when objects are irregular (non-cubic) and obscuring each other. For example, the text prompt ``an astronaut riding a horse'' can not be represented by two non-overlapping bounding boxes. To avoid these limitations while still achieving object decomposition, 
we propose \emph{GraphDreamer}, which takes a scene graph (\eg, \cite{krishna2016visual}) as input and generates a compositional 3D scene. Unlike 3D bounding boxes, scene graphs are spatially more relaxed and can model complex object interaction. While scene graphs are generally 
easier to specify than spatial 3D layouts, we also
design a text prompt to query ChatGPT that enables the automatic generation of a scene graph from unstructured text. See Figure~\ref{fig:teaser} for an illustrative example.

GraphDreamer is guided by the insight that a scene graph can be decomposed into a separate and semantically unambiguous text description of every node and edge\footnote{Cf.\ the assumption of {\em independent causal mechanisms}~\citep{Scholkopfetal21}}. The decomposition of a scene graph into multiple textual descriptions makes it possible to distill knowledge from text-to-image diffusion models, similar to common text-to-3D methods. Specifically, to allow each object to be disentangled from the other objects in the scene, we use separate identity-aware positional encoder networks (\ie, object feature fields) to encode object-level semantic information and a shared {\em Signed Distance Field}~(SDF) network to decode the SDF value from identity-aware positional features. The color value is decoded in a way similar to the SDF value. 
Scene-level rendering is performed by integrating objects based on the smallest SDF value at each sampled point in 3D space. 
More importantly, with both SDF and color values of each object, we propose an identity-aware object rendering that, in addition to a global rendering of the entire 3D scene, renders different objects separately. Our local identity-aware rendering allows the gradient from the text-dependent distillation loss (\eg, score distillation sampling~\cite{poole2022dreamfusion}) to be back-propagated selectively to corresponding objects without affecting the other objects. 
The overall 3D scene will be simultaneously optimized with the global text description to match the scene semantics to the global text. 
In summary, we make the following contributions:

\begin{itemize}
    \item To the best of our knowledge, GraphDreamer is the first 3D generation method that can synthesize compositional 3D scenes from either scene graphs or unstructured text descriptions. No 3D bounding boxes are required as input.\vspace{.75mm}
    \item GraphDreamer uses scene graphs to construct a disentangled 
    representation where each object is optimized via its related text description, avoiding object-level ambiguity.
    \vspace{.75mm}
    \item GraphDreamer is able to produce high-fidelity complex 3D scenes with disentangled objects, outperforming both state-of-the-art text-to-3D methods and existing 3D-bounding-box-based compositional text-to-3D methods.
    \vspace{.75mm}
    \item In Appendix~\ref{app:is}, we envision a new paradigm of semantic 3D reconstruction -- \emph{Inverse Semantics}, where a vision-language model (\eg, GPT4-V) is used to extract a scene graph from an input image (\ie, scene graph encoder) and GraphDreamer is used to generate a compositional 3D scene from the scene graph (\ie, scene graph decoder).
\end{itemize}

%% file: sec/2_related.tex
\section{Related Work}

\textbf{Text-to-2D generation}.
Driven by large-scale image-text aligned datasets~\cite{schuhmann2022laion5b}, text-to-image generation models~\cite{balaji2023ediffi,saharia2022photorealistic, rombach2021highresolution} have made great progress in producing highly realistic images. 
Among these models, generative diffusion models learn to gradually transform a noisy latent $z$ with noise $\epsilon$ typically from a Gaussian distribution, towards image data $x$ that reproduce the semantics of a given text prompt $y$. 
This generative process slowly adds structure to the noise, based on a weighted denoising score matching objective \cite{Nichol2021ARXIV, kingma2023variational}. 

\vspace{.9mm}
\noindent\textbf{2D-lifting for 3D generation}.
In contrast to existing text-to-image generation models, text-guided 3D generative models~\cite{Jain2022CVPR, Mohammad_Khalid_2022, poole2022dreamfusion, wang2022score, metzer2022latentnerf, tsalicoglou2023textmesh, text2mesh2022, nam20223dldm} usually optimize a 3D model by guiding its randomly rendered 2D view based on the pretraining knowledge of some text-to-image generation model, because of the shortage of text-3D paired assets.  
DreamFusion~\cite{poole2022dreamfusion} and subsequent work~\cite{lin2023magic3d, wang2023prolificdreamer, chen2023fantasia3d, xu2023dream3d,shi2023MVDream} propose to optimize the 3D model by distilling a pretrained diffusion model~\cite{saharia2022photorealistic, rombach2021highresolution} via score distillation sampling~\cite{wang2022score}.

\vspace{.9mm}
\noindent\textbf{Generate objects with SDF}. 
In text-to-3D generation, recent works~\cite{Jain2022CVPR, poole2022dreamfusion,wang2023score, lin2023magic3d, wang2023prolificdreamer, wang2022score, liu2023zero, metzer2022latentnerf, shi2023MVDream} parameterized the 3D scene as a NeRF~\cite{Mildenhall2020ECCV, Barron2021ICCV} or a hybrid pipeline combining NeRFs with a mesh refiner~\cite{jain2022zeroshot, lin2023magic3d, shen2021dmtet, tsalicoglou2023textmesh}. 
In our approach, we use a signed distance field (SDF) as the geometry representation instead of NeRF densities, as we aim at modeling multi-object scenes in a compositional way, where objects may be coupled in various ways. 
SDF provides crucial inside/outside information, allowing for geometry constraints to prevent unexpected intersections between objects, and is ideal for complex scenes as it facilitates customization of initial locations and scales of object SDFs.

\vspace{.9mm}
\noindent\textbf{Hybrid 3D representation for disentanglement}. Another line of work uses hybrid representations to learn disentangled 3D objects~\cite{li2022eyenerf,xu2022hybrid,liu2022structural}. The works \cite{feng2022capturing,feng2023learning} put forward a hybrid approach that represents the face/body as meshes and the hair/clothing as NeRFs, enabling a disentangled reconstruction of avatars. \cite{zhang2023text} adopts this representation and proposes a text-to-3D method that generates compositional head avatars. However, the use of a parametric head model limits this method to human head generation. Their disentanglement only applies to two objects (\eg, face and hair), and in contrast, ours can be used for multiple objects.

%% file: sec/3_preliminary.tex
\section{Preliminaries}

\textbf{Score distillation sampling (SDS)}. SDS~\cite{poole2022dreamfusion} is a technique that optimizes a 3D model by distilling a pretrained text-to-image diffusion model. Given a noisy image $z_t$ rendered from a 3D model parameterized by $\Theta$, and a pretrained text-to-image diffusion model with a learned noise prediction network $\epsilon_{\phi}(\cdot)$, 
SDS uses $\epsilon_{\phi}(\cdot)$ as a score function that predicts the sampled noise $\epsilon$ contained in $x_t$ at noise level $t$ as $\hat{\epsilon}_\phi(y,t) = \epsilon_\phi(z_t; y,t)$, where $y$ is a given conditional text embedding. 
The score is then used to warp the noisy $z_t$ towards real image distributions, by guiding the direction of the gradients that update the parameters $\Theta$ of the 3D model:  
\begin{equation}
\label{eq:sds}
    \nabla_\Theta\mathcal{L}(z; y)
        = \mathbb{E}_{t,\epsilon}\biggl[w(t)\Bigl(\hat{\epsilon}_\phi(y,t) - \epsilon\Bigr)\frac{\partial z}{\partial \Theta}\biggr] 
\end{equation}
where $w(t)$ is a weighting function that depends on $t$ and $\epsilon$ is the sampled isotropic Gaussian noise, $\epsilon \sim \mathcal{N}(\mathbf{0}, \mathbf{I})$.

\vspace{1mm}
\noindent\textbf{SDF volume rendering}.
To render a pixel color of a target camera, we cast a ray $\br$ from the camera center $\bo$ along its viewing direction $\bd$, then sample a series of points $\bp = \bo + t\bd$ in between the near and far intervals $[t_n,t_f]$. 
Following NeRF~\cite{Mildenhall2020ECCV}, the ray color $C(\br)$ can be approximated by integrating the point samples,
\begin{equation}
\label{eq:volume_rendering}
 C(\br) = \sum_{i=1}^{N} w_i c_i = \sum_{i=1}^{N} T_i \alpha_i c_i
\end{equation}
where $w_i$ is the color weighting function, $T_i$ represents the cumulative transmittance, which is calculated as $T_i=\prod_{j=1}^{i-1} (1-\alpha_j)$, and $\alpha_i$ denotes the piece-wise constant opacity value for the $i$-th sub-interval, with $\alpha_i \in [0,1]$.

Unlike NeRFs that directly predict the density for a given position $\bp$, methods based on implicit surface representation learn to map $\bp$ to a signed distance value with a trainable network and extract the density~\cite{Yariv2021NEURIPS} or opacity~\cite{Wang2021NEURIPSa} from the SDF with a deterministic transformation. 
We extract the opacity following NeuS~\cite{Wang2021NEURIPSa}. NeuS formulates the transformation function based on an unbiased weighting function, which ensures that the pixel color is dominated by the intersection point of camera ray with the zero-level set of SDF, 
\begin{equation}
 \alpha_i
 = \text{max} \left( \frac{\Phi_\beta(u_i) - \Phi_\beta(u_{i+1})}{\Phi_\beta(u_i)} , 0\right)
 \label{eq:neus}
\end{equation}
where $\Phi_\beta(\cdot)$ is the Sigmoid function with a trainable steepness $\beta$, and $u_i$ is the SDF value of the sampled position.

%% file: sec/4_method.tex
\section{Method}

Consider generating a scene of $M$ objects, $\mathcal{O}=\{o_i\}_{i=1}^M$ from a global text prompt $y^g$. 
When the scene is complex or has many attributes and inter-object relationships to specify, $y^g$ will become very long, and the generation will be accompanied by guidance collapse~ \cite{chefer2023attendandexcite, li2023divide}. 
We thus propose to first generate a \textbf{scene graph} $\mathcal{G}(\mathcal{O})$ from $y^g$ following the setting of~\cite{krishna2016visual}, which precisely describes object attributes and inter-object relationships. We provide an example of a four-object scene in Figure~\ref{fig:teaser} for better illustration. 
\input{figures/pipeline.tex}

\subsection{Leveraging Scene Graphs for Text Grounding}
Given user text input $y^g$, objects $\{o_i\}_{i=1}^M$ in the text (which can be detected either manually or automatically, \eg, using ChatGPT\footnote{ChatGPT4, \href{https://chat.openai.com}{https://chat.openai.com}}) form the \textcolor{myred1}{\textbf{nodes}} in graph $\mathcal{G}(\mathcal{O})$, as shown in Figure~\ref{fig:teaser}. To provide more details to an object $o_i$, the user can add additional descriptions, such as ``Wise-looking'' and ``Leather-bound'', which become the \textcolor{mygreen1}{\textbf{attributes}} attached to $o_i$ in $\mathcal{G}(\mathcal{O})$. 
Combining $o_i$ with all its attributes simply by commas, we get an \textcolor{myred1}{\textbf{object prompt} $y^{(i)}$} for $o_i$ that can be processed by text encoders. 

The relationship between each pair of objects $o_i$ and $o_j$ is transformed into \textcolor{myblue1}{\textbf{edge}} $e_{i,j}$ in $\mathcal{G}(\mathcal{O})$. 
For instance, the edge between node ``Wizard'' and ``Desk'' is ``Standing in front of''. For a graph with $M$ nodes, there are possibly $C_2^M$ edges. By combining $o_i$, $e_{i,j}$, and $o_j$, we obtain \textcolor{myblue1}{\textbf{edge prompt} $y^{(i,j)}$} that exactly defines the pairwise relationship, \eg, ``Wizard standing in front of Wooden Desk''. Note that there might be no edge between two nodes, \eg, between ``Wizard'' and ``Stack of Ancient Spell Books''. We denote the number of existing edges in $\mathcal{G}(\mathcal{O})$ as $K$, with $K \leq C_2^M$. 

From this example, we also see that using graph $\mathcal{G}(\mathcal{O})$ is a better way to customize a scene compared to a pure text description $y^g$, in terms of both flexibility in attaching attributes to objects and accuracy in defining relationships. 
By processing the input scene graph, we now obtain a set of $(1+M+K)$ prompts $\mathcal{Y}\bigl(\mathcal{G}(\mathcal{O})\bigr)$ as:
\begin{equation}
\label{eq:graph_prompts}
    \mathcal{Y}\bigl(\mathcal{G}(\mathcal{O})\bigr)=\bigl\{y^g, y^{(i)}, y^{(i,j)} \mid o_i \in \mathcal{O}, e_{i,j} \in \mathcal{G}(\mathcal{O}) \bigr\}
\end{equation}
which are used to guide scene generation from the perspective of both individual objects and pairwise relationships.

GraphDreamer consists of three learnable modules: a positional feature encoder $\mathcal{F}_{\theta}(\cdot)$, a signed distance network $u_{\phi_1}(\cdot)$, and a radiance network $c_{\phi_2}(\cdot)$. The entire model is parameterized by $\Theta=\{\theta, \phi_1, \phi_2\}$. There are \textbf{two goals} in optimizing GraphDreamer: (i) to model the complete geometry and appearance of each object, and (ii) to ensure that object attributes and interrelationships are in accordance with the scene graph $\mathcal{G}(\mathcal{O})$. 
The overall training process is illustrated in Figure \ref{fig:pipeline}. 

\subsection{Disentangled Modeling of Objects}
\label{subsec:obj_rendering}
Positional encodings are useful for networks to identify the location it is currently processing. 
To achieve the first goal of making objects separable, we need to additionally identify which object a position belongs to. Therefore, instead of one positional feature embedding, we encode a position $\bp$ into multiple feature embeddings by introducing a set of positional hash feature encoders, each parameterized by $\theta_i$, corresponding to the number of objects,
\begin{equation}
    \mathcal{F}_{\theta}(\cdot)=\{\mathcal{F}_{\theta_i}(\cdot)\}_{i=1}^M ~~~~~ \theta=\{\theta_i\}_{i=1}^M
\end{equation}
These feature encoders then form different object fields, \ie, one field per object across the same scene space $\Omega \subset \mathbb{R}^3$. 

\vspace{1mm}
\noindent\textbf{Individualized object fields}. Given a position $\bp \in \Omega$, the feature that forms the field of object $o_i$ is obtained as: 
\begin{equation} 
    f^{(i)}(\bp) = \mathcal{F}_{\theta_i}(\bp) \in \mathbb{R}^F ~~~~ i \in \{1, \cdots, M\} 
\end{equation}
where $F$ is the number of feature dimensions, the same for all $\mathcal{F}_{\theta_i}(\cdot)$. Here, for each $\mathcal{F}_{\theta_i}(\cdot)$ we adopt the multi-resolution hash grid encoding from Instant NGP~\cite{Mueller2022SIGGRAPH} following~\cite{lin2023magic3d,wang2023prolificdreamer,shi2023MVDream} to reduce computational cost.

These identity-aware feature embeddings are then passed to the shared shallow MLPs for SDF and color prediction, \eg, the SDF $u^{(i)}(\bp) \in \mathbb{R}$ and color $c^{(i)}(\bp) \in \mathbb{R}^3$ values for object $o_i$'s field are predicted as:
\begin{equation}
    u^{(i)}(\bp) = u_{\phi_1}\bigl(f^{(i)}(\bp)\bigr) ~~~~~~
    c^{(i)}(\bp) = c_{\phi_2}\bigl(f^{(i)}(\bp)\bigr)    
\end{equation}
where $u^{(i)}(\bp)$ indicates the signed distance value from position $\bp$ to the closest surface of object $o_i$, with negative values inside $o_i$ and positive values outside, and $c^{(i)}(\bp)$ the color value in $o_i$'s field where only object $o_i$ is considered. Here, we follow prior work~\cite{Yariv2020NIPS} to initialize the SDF approximately as a sphere. We transform $u^{(i)}(\bp)$ into opacity with Eq.~(\ref{eq:neus}) as $\gamma^{(i)}(\bp)$ for the volume rendering of object fields. 

In scenes where mutual-obscuring relationships are involved, to generate the complete geometry and appearance for each object, we need to make hidden object surfaces visible. Therefore, the scene needs to be properly decomposed before rendering the objects.

\vspace{1mm}
\noindent\textbf{Scene space decomposition}. Intuitively, a position $\bp \in \Omega$ can be occupied by at most one object.  
Since the SDF determines the boundary of an object, $\bp$ can thus be identified as belonging to object $o_i$ if its SDF values $\{u^{(i)}\}_{i=1}^M$ are minimized at index $i$. 
Based on this, we define an one-hot identity (column) vector $\boldsymbol{\lambda}(\bp)$ for each position $\bp$ as:  
\begin{equation}
\label{eq:soft_identity}
    \boldsymbol{\lambda}(\bp) = \argmax_{i=1, \cdots, M}\bigl\{ -u^{(i)}(\bp)\bigr\} \in \{0, 1\}^M 
\end{equation}
Based on $\boldsymbol{\lambda}(\bp)$, we can decompose the scene into identity-aware sub-spaces and render each object individually with all other objects removed.

\vspace{1mm}
\noindent\textbf{Identity-aware object rendering}. 
To render object $o_i$, given a position $\bp$, we multiply its opacity $\gamma^{(i)}$ in $o_i$'s field with $\lambda^{(i)}(\bp)$ to obtain the opacity for only object $o_i$ as:
\begin{equation}
    \label{eq:semantic_opacity}
    \alpha^{(i)}(\bp) = \lambda^{(i)}(\bp) \cdot \gamma^{(i)} \in [0, +\infty) 
\end{equation}
where $\lambda^{(i)}(\bp)$ is the $i$-th element in vector $\boldsymbol{\lambda}(\bp)$.
If $\lambda^{(i)}(\bp) = 1$, which means $\bp$ is identified as most likely to be occupied by object $o_i$, we have opacity $\alpha^{(i)}(\bp) \geq 0$ in object $o$'s field only, while in all other object fields $\bp$ will be empty. Based on this identity-aware opacity $\alpha^{(i)}(\bp)$, we can obtain the ray color with object $o_i$ present only as: 

\vspace{-2mm}
\begin{equation} 
\label{eq:render}
    C_{\br}^{(i)} = \sum_\bp \alpha^{(i)}(\bp)T^{(i)}(\bp)\cdot c^{(i)}(\bp) 
\end{equation}
\vspace{-2.25mm}

\noindent where the cumulative transmittance $T^{(i)}(\bp_j)$ is defined following Eq.~(\ref{eq:volume_rendering}). 
By aggregating all rendered pixels, an object image $C^{(i)}$ is obtained, which contains object $o_i$ only. 
With $C^{(i)}$ and the object prompt $y^{(i)}$ from the scene graph $\mathcal{G}(\mathcal{O})$, we can thus define the object SDS loss following Eq.~(\ref{eq:sds}) as:
\begin{equation}
\label{eq:obj_sds}
    \nabla_\Theta\mathcal{L}^{(i)}\bigl(C^{(i)}; y^{(i)}\bigr) ~~~~ o_i \in \mathcal{O}
\end{equation}
to supervise GraphDreamer at the object level.

\subsection{Pairwise Modeling of Object Relationships}
\label{subsec:edge_rendering} 
By building up a set of identity-aware object fields, we are now able to render objects in $\mathcal{G}(\mathcal{O})$ individually to match $y^{(i)}$. 
To make two related objects $o_i$ and $o_j$ respect the relationship defined in edge prompt $y^{(i,j)}$, we need to render $o_i$ and $o_j$ jointly. Therefore, a combination of two object fields is required. 

\vspace{1mm}
\noindent\textbf{Edge rendering}.
Given an edge $e_{i,j}$ connecting nodes $o_i$ and $o_j$, we can accumulate an edge-wise opacity $\alpha^{(i,j)}(\bp)$ at position $\bp$ from opacity values of $o_i$ and $o_j$ based on the one-hot identity vector $\boldsymbol{\lambda}(\bp)$ as: 
\begin{equation} 
        \alpha^{(i,j)}(\bp) = \sum_{k\in\{i,j\}} \lambda^{(k)}(\bp) \cdot \gamma^{(k)}(\bp) 
\end{equation}
and similarly, an edge-wise color $c^{(i,j)}(\bp)$ at position $\bp$ as:
\begin{equation}
    c^{(i,j)}(\bp) = \sum_{k\in\{i,j\}} \lambda^{(k)}(\bp) \cdot c^{(k)}(\bp)   
\end{equation}
from which we can render a ray color across two object fields of $o_i$ and $o_j$ following the same integration process as Eq.~(\ref{eq:volume_rendering}), and obtain an edge image $C^{(i,j)}$ in with both objects involved. We can thus define an edge SDS loss as:
\begin{equation}
\label{eq:edge_sds}
    \nabla_\Theta\mathcal{L}^{(i,j)}\bigl(C^{(i,j)}; y^{(i,j)}\bigr) ~~~~ e_{i,j} \in \mathcal{G}(\mathcal{O})  
\end{equation}
to match the edge prompt $y^{(i,j)}$.

\vspace{1mm}
\noindent\textbf{Scene rendering}.
To provide global scene graph $\mathcal{G}(\mathcal{O})$ constraints, we further render the whole scene globally by combining all object fields together. 
Similarly as for the edge rendering, we use $\boldsymbol{\lambda}(\bp)$ to accumulated the global opacity $\alpha^g(\bp)$ and a global color $c(\bp)$ at position $\bp$ over all object fields as: 
\begin{equation}
\label{eq:global_opacity} 
    \alpha^g(\bp) = \boldsymbol{\lambda}^T(\bp) \cdot \boldsymbol{\gamma}(\bp)~~~~~~c^g(\bp) = \boldsymbol{\lambda}^T(\bp) \cdot \boldsymbol{c}(\bp) 
\end{equation}
with $\boldsymbol{\gamma}(\bp)$ and $\boldsymbol{c}(\bp)$ denote the column vectors of opacity and color values at $\bp$ over all object fields. Through the integration process, we can render the entire scene into a scene image $C^g$ that represents the entire scene graph. We define a scene-level SDS loss, $\nabla_\Theta\mathcal{L}^g\bigl(C^g; y^g\bigr)$, 
to globally optimize GraphDreamer to match the scene prompt $y^g$.

\vspace{1mm}
\noindent\textbf{Efficient SDS guidance}.
So far, a number of $(M + K + 1)$ SDS losses are introduced, corresponding to the number of prompts we obtained in Eq.~(\ref{eq:graph_prompts}). 
However, optimizing all constraints jointly is intractable. Instead, in each training step, we include two SDS losses only: 
\textbf{(i)} an object SDS loss Eq.~(\ref{eq:obj_sds}) for only one object $o_i$, with each step choosing a different object $o_i$ looping through $\mathcal{O}$; 
\textbf{(ii)} an edge SDS loss Eq.~(\ref{eq:edge_sds}) of one $e_{i,j}$ connected to $o_i$, looping through all existing edges connected to $o_i$. 
The scene SDS loss is included only in the training step after each traversal of $\mathcal{O}$, and is used solely in that step without any other SDS loss. 
Thus, the total SDS loss for optimizing GraphDreamer is: 
\begin{equation}
\label{eq:total_sds}
    \mathcal{L}_{SDS} = \begin{cases}
        \nabla_\Theta\mathcal{L}^{(i)} + \nabla_\Theta\mathcal{L}^e, & i = s\text{ \% } (M+1) \\
        \nabla_\Theta\mathcal{L}^g, & \text{others} 
    \end{cases}
\end{equation}
where $s$ indexes the current training step, and $e$ refers to one of the edges connected with $o_i$. 

\subsection{Training Objectives}
\label{sec_training}
\input{figures/baseline_compare}

Apart from the SDS guidance, to further optimize the prediction of unobserved positions, \ie, inside objects and on hidden surfaces at object intersections, we include two geometry constraints for physically plausible shape completion. 

\vspace{1mm}
\noindent\textbf{Penetration constraint}. 
Since each point $\bp \in \Omega$ can be inside or on the surface of at most one object $o$ in the set of objects $\mathcal{O}$, and for regions outside the actual objects, $u(\bp) \in (0, +\infty)$ with $\bp \in \Omega \setminus \mathcal{O}$, we define a penetration measurement at point p:
\begin{equation}
        \mathcal{N}^-(\bp) = \sum_{i=1}^M \max \Bigl\{ \text{sgn}\bigl(-u^{(i)}(\bp)\bigr), 0 \Bigr\}    
\end{equation}
where $\text{sgn}(x)$ is the sign function. Intuitively, $\mathcal{N}^-(\bp)$ measures the number of objects inside which the point $\bp$ is located, according to the predicted $u^{(i)}(\bp)$. 
Using this measurement, we propose to implement a penetration constraint, 
\begin{equation}
\label{eq:penet_loss}
    \mathcal{L}_{penet}(\bp) = \max \bigl\{0, \mathcal{N}^-(\bp) - 1\bigr\}
\end{equation}
to constrains the penetration number $\mathcal{N}^-(\bp)$ not to exceed $1$, which is averaged over all sampled points during training.

\vspace{1mm}
\noindent\textbf{Eikonal constraint}. 
At each sampled position $\bp$, we adopt the Eikonal loss~\cite{Gropp2020ICML} on SDF values $u^{(i)}(\bp)$ from all object fields, formulated as
\begin{equation}
    \mathcal{L}_{eknl} = \frac{1}{NM} \sum_{i, p}\biggl( \Big\| \nabla u^{(i)}(\bp) \Big\|_2 - 1 \biggr)
\end{equation}
where $N$ is the size of the sample set $\mathcal{P}_\br$ on ray $\br$.

\vspace{1mm}
\noindent\textbf{Total training loss}.
Our final loss function for training GraphDreamer thus consists of four terms: 
\begin{equation}
\label{eq:total_loss}
    \mathcal{L}_{\Theta} = \beta_1 \mathcal{L}_{SDS} + \beta_2 \mathcal{L}_{penet} + \beta_3 \mathcal{L}_{eknl} 
\end{equation}
where $\{\beta_1, \beta_2,\beta_3\}$ are hyperparameters. 

%% file: figures/pipeline.tex
\begin{figure*}[t]
    \centering
    \vspace{-2mm}
    \includegraphics[width=\linewidth]{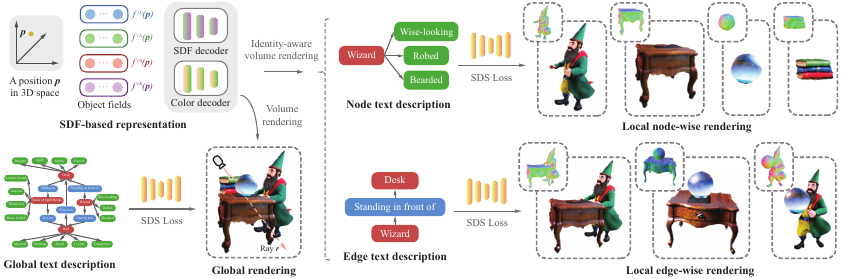}
    \vspace{-5mm}
    \caption{\footnotesize The overall pipeline of GraphDreamer. Specifically, GraphDreamer first  decomposes the scene graph into global, node-wise and edge-wise text description, and then optimizes the SDF-based objects in the 3D scene using their corresponding text description.}
    \label{fig:pipeline}
    \vspace{-2mm}
\end{figure*}

%% file: figures/baseline_compare.tex
\begin{figure*}[t]
    \centering
    \vspace{-0.4cm}
    \includegraphics[width=\linewidth]{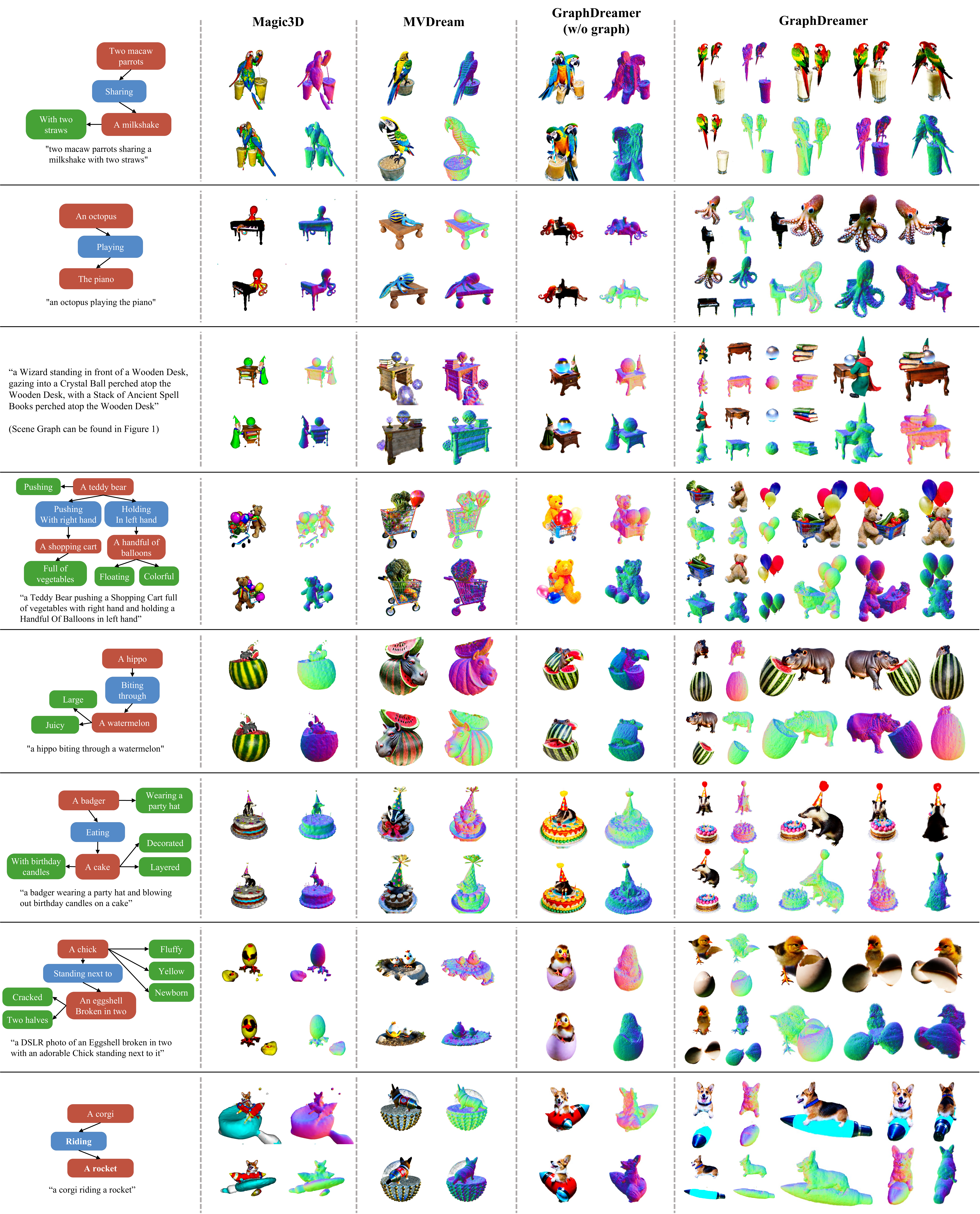}
\caption{\footnotesize Qualitative comparison with baseline approaches and the ablated configuration (w/o graph). GraphDreamer generates scenes with all composing objects being separable. Moreover, with accurate guidance from scene graphs, object attributes and inter-object relationships produced by GraphDreamer match the given prompts better. We recommend to zoom-in for details. }
\label{fig:baseline_compare}
\vspace{-0.4cm}
\end{figure*}

%% file: sec/5_experiments.tex
\input{00tables/baseline}
\section{Experiments and Results} 
\label{sec:experiments}

\textbf{Implementation details}. 
We adopt a two-stage coarse-to-fine optimization strategy following previous work~\cite{poole2022dreamfusion, chen2023fantasia3d, lin2023magic3d, shi2023MVDream}.   
In the first stage of $10$K denoising steps, we render images of $64 \times 64$ resolution only for faster convergence and use DeepFloyd-IF\footnote{DeepFloyd-IF, \href{https://huggingface.co/DeepFloyd/IF-I-XL-v1.0}{huggingface.co/DeepFloyd/IF-I-XL-v1.0}}~\cite{deepfloyd2023if, saharia2022photorealistic} as our guidance model, which is also trained to generate $64$px images. 
In the second stage of $10$K steps, the model is refined by rendering 256px images and uses Stable Diffusion\footnote{Stable diffusion, \href{https://huggingface.co/stabilityai/stable-diffusion-2-1-base}{huggingface.co/stabilityai/stable-diffusion-2-1-base}}~\cite{rombach2021highresolution} as guidance. 
Both stages of optimization use $1$ Nvidia Quadro RTX 6000 GPU; GraphDreamer uses 16.88/18.58/20.05 GB for generating $2$/$3$/$4$ objects, while Magic3D/MVDream uses 11.44/20.33 GB.

\vspace{1mm}
\noindent\textbf{Baseline approaches}. 
We report results of two state-of-the-art approaches, Magic3D~\cite{lin2023magic3d} and MVDream~\cite{shi2023MVDream}. Both approaches use a frozen guidance model without additional learnable module~\cite{wang2023prolificdreamer} and do not have special initialization requirements~\cite{chen2023fantasia3d} for geometry.
We use the same guidance models and strategy to train Magic3D, 
while for MVDream, since it proposes to use a fine-tuned multi-view diffusion model\footnote{MVDream-sd-v2.1-base-4view, \href{https://huggingface.co/MVDream/MVDream}{huggingface.co/MVDream/MVDream}}, we follow its official training protocol and use its released diffusion model as guidance. 
The experimental results of GraphDreamer and the baselines are all obtained after training for 20K steps in total.

\vspace{1mm}
\noindent\textbf{Evaluation criteria}.
We report the CLIP Score~\cite{Radford2021ARXIV} in quantitative comparison with baseline models and evaluation on object decomposition. The metric is defined as:  
\begin{equation}
    \text{CLIPScore}(C, y) = \cos{\bigl<E_C(C), E_Y(y)\bigr>}
\end{equation}
which measures the similarity between a prompt text $y$ for an image $C$ and the actual content of the image, with $E_C(C)$ the visual CLIP embedding and $E_Y(y)$ the textual CLIP embedding,   both encoded by the same CLIP model\footnote{CLIP B/32, \href{https://huggingface.co/openai/clip-vit-base-patch32}{huggingface.co/openai/clip-vit-base-patch32}}. 

\subsection{Comparison to Other Methods}
\input{00tables/analysis_decompo}

We report quantitative results in Table~\ref{tab:baseline_compare} and qualitative results in Figure~\ref{fig:baseline_compare}. 
The figures in Table~\ref{tab:baseline_compare} are summarized from CLIP scores of $30$ multi-object scenes, with the number of objects $\geq 2$. 
GraphDreamer (full model) achieves the highest CLIP score. 
Qualitative results shown in Figure~\ref{fig:baseline_compare} also suggest that GraphDreamer is generally more applicable in producing multiple objects in a scene with various inter-object relationships.  
As can be observed from baseline results, semantic mismatches are more commonly found, such as, in the example of ``two macaw parrots sharing a milkshake with two straws'', Magic3D generates two ``milkshakes'' and MVDream produces one ``parrot'' only, which mixed up the number attribute of ``milkshake'' and ``parrots'', and in the example of "a hippo biting through a watermelon", two objects ``hippo'' and ``watermelon'' are blended into one object;
GraphDreamer, on the other hand, models both individual attributes and inter-object relationships correctly based on the guidance of the input scene graph.

\subsection{Ablation Study} 
\input{figures/more_examples}

To evaluate the effect of introducing scene graph $\mathcal{G}(\mathcal{O})$ as guidance on preventing guidance collapse, we consider to drop $\mathcal{G}(\mathcal{O})$ from GraphDreamer, termed {\textbf{w/o graph}, and compare the results in Table~\ref{tab:baseline_compare} and Figure~\ref{fig:baseline_compare}. 
GraphDreamer (w/o graph) collapses the conditioning text from a set of prompts defined in Eq.~(\ref{eq:graph_prompts}) to only the global prompt $y^g$ and thus involves no object/edge rendering 
(described in Section~\ref{subsec:obj_rendering} and \ref{subsec:edge_rendering}) 
in training; $L_{SDS}$ from Eq.~(\ref{eq:total_sds}) becomes $\nabla_\Theta\mathcal{L}^g$, with all other implementation details unchanged. 
As reported in the fourth column of Table~\ref{tab:baseline_compare}, the mean CLIP score of the ablated configuration decreases by more than $3$ std. compared to the full model, indicating that the ability of this configuration to generate 3D scenes that match given prompts is significantly reduced. 
The results in Figure~\ref{fig:baseline_compare} also corroborate such a decline, given that the problems such as missing objects and attribute confusion arise again. 
Both evaluations suggest the need for the scene graph $\mathcal{G}(\mathcal{O})$ in GraphDreamer for combating guidance collapse problems.

\subsection{Decomposition Analysis}
\label{subsec:analy_decompo}

We consider scenes with the number of objects $\geq 2$. 
To further quantitatively evaluate whether objects in a scene are well separated and rendered into object images individually, 
we calculate two CLIP metrics for each object image $C^{(i)}$: \textcolor{myred1}{\textbf{(i) w.\ Self Prompt}} (abbr., wSP) refers to the CLIP score between $C^{(i)}$ and its own object prompt $y^{(i)}$, and \textcolor{myblue1}{\textbf{(ii) w.\ Other Prompts}} (abbr., wOP) refers to the CLIP scores between $C^{(i)}$ and all other object prompts $\{y^{(j)}, j\neq i, o_j \in \mathcal{O}\}$ in the same scene.
Intuitively, if a scene is well decomposed, each object image $C^{(i)}$ should contain one object $o_i$ only without any part of other objects, and thus the scores wSP should be much higher than the scores wOP.  
Table~\ref{subtab:decompo} reports a statistical summary on the metrics of $64$ objects, for each object, we render images from $4$ orthogonal views $C^{(i)}_v$ ($v=1,2,3,4$), and thus we get $4$ wSP scores and multiple wOP scores per object. 
We calculate the mean and standard deviation (std.) of these wSP and wOP scores separately over the view images. The figures reported in the table are averaged over all $64$ objects. 
The mean and std. values are also presented in an error band graph in Figure~\ref{subfig:decompo}, where the $x$-axis is the index of objects.  
From this graph it can be observed more obviously that the wSP CLIP score is significantly higher than the wOP CLIP score, without overlap between the mean wSP scores and the $ \text{mean} \pm 3~\text{std.}$ error band of the wOP scores, which shows that the scenes are properly decomposed into individual objects. 
More compositional examples can be found in Figure~\ref{fig:more_examples}. 

%% file: 00tables/baseline.tex
\begin{table}[!t]
\setlength\tabcolsep{2.2pt}
\renewcommand{\arraystretch}{1.2}
\footnotesize
    \begin{tabular}{c|cccc} 
        \begin{tabular}[c]{@{}c@{}}CLIP\\ Score\end{tabular} & Magic3D~\cite{lin2023magic3d} & MVDream~\cite{shi2023MVDream} & \begin{tabular}[c]{@{}c@{}}GraphDreamer\\(w/o graph)\end{tabular} &\cellcolor{Gray} GraphDreamer\\ \shline
        mean & 0.3267 & 0.3102 & 0.3019 &\cellcolor{Gray} \textbf{0.3326} \\
        std. & 0.0362 & 0.0061 & 0.0254 & \cellcolor{Gray}0.0252 \\
    \end{tabular}
\caption{\footnotesize Quantitative results. The mean and standard deviation (std.) values are summarized from CLIP scores of $30$ multi-object scenes, with the number of objects in the scene $\geq 2$. For better comparison, we provide the result of GraphDreamer (w/o graph), the configuration with the scene graph $\mathcal{G}(\mathcal{O})$ dropped from GraphDreamer, and thus the conditioning text for all renderings collapse to a single $y^g$. }
\label{tab:baseline_compare}
\vspace{-2mm}
\end{table}

%% file: 00tables/analysis_decompo.tex
\begin{figure}[tb]
\centering
    \begin{minipage}{\linewidth}
        \setlength\tabcolsep{10pt}
\renewcommand{\arraystretch}{1.3}
        \centering
        \footnotesize
            \begin{tabular}{c|ccccc} 
                \multirow{2}{*}{\begin{tabular}[c]{@{}l@{}}CLIP Score\end{tabular}} & \multicolumn{2}{c}{w. Self Prompt $\uparrow$} &  \multicolumn{2}{c}{w. Other Prompts $\downarrow$} \\  
                 & mean & std. &  mean & std. \\ \shline
                GraphDreamer & 0.308 & 0.012 & 0.201 & 0.009 \\ 
            \end{tabular}
        \tabcaption{\footnotesize The CLIP scores of individual object images $C^{(i)}$. Metric \textcolor{myred1}{\textbf{w. Self Prompt}} refers to scores calculated between $C^{(i)}$ and its own prompt $y^{(i)}$, and \textcolor{myblue1}{\textbf{w. Other Prompts}} between $C^{(i)}$ and prompts of all other objects in the same scene $\{y^{(j)}, j\neq i, o_j \in \mathcal{O}\}$. Detailed experimental settings and analysis on these figures as well as the on the chart showing in Figure~\ref{subfig:decompo}, can be found in Subsection~\ref{subsec:analy_decompo}. }
        \label{subtab:decompo}
    \end{minipage}
    \\
    \begin{minipage}{\linewidth}
        \centering
        \includegraphics[width = \linewidth]{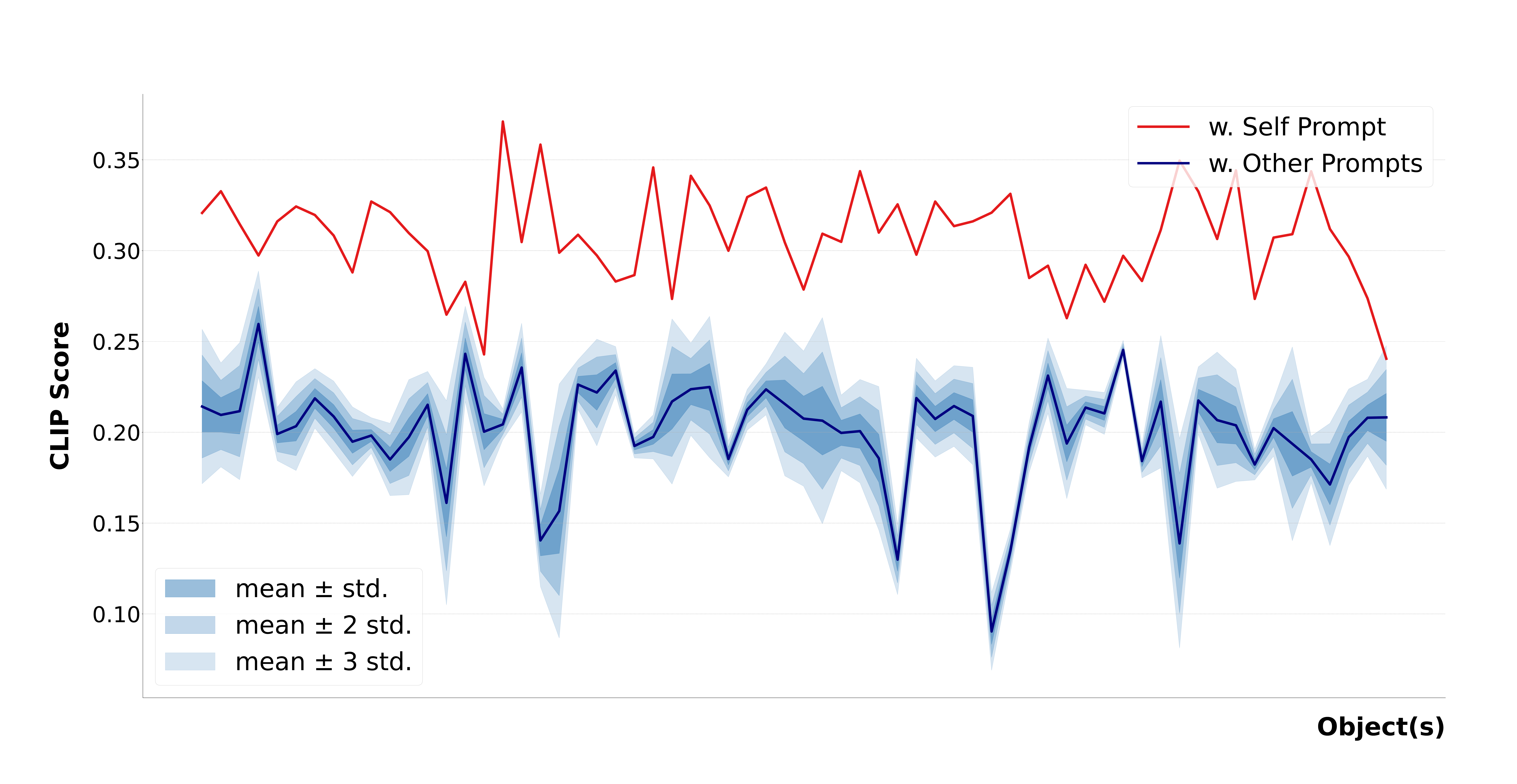}
        \vspace{-9mm}
        \figcaption{\footnotesize Error bands of object CLIP scores.} 
        \label{subfig:decompo}
    \end{minipage}
\vspace{-0.3cm}
\end{figure}
\captionsetup[figure]{name=Figure}

%% file: figures/more_examples.tex
\begin{figure*}[t!]
    \centering
    \vspace{-3mm}
    \includegraphics[width=.97\linewidth]{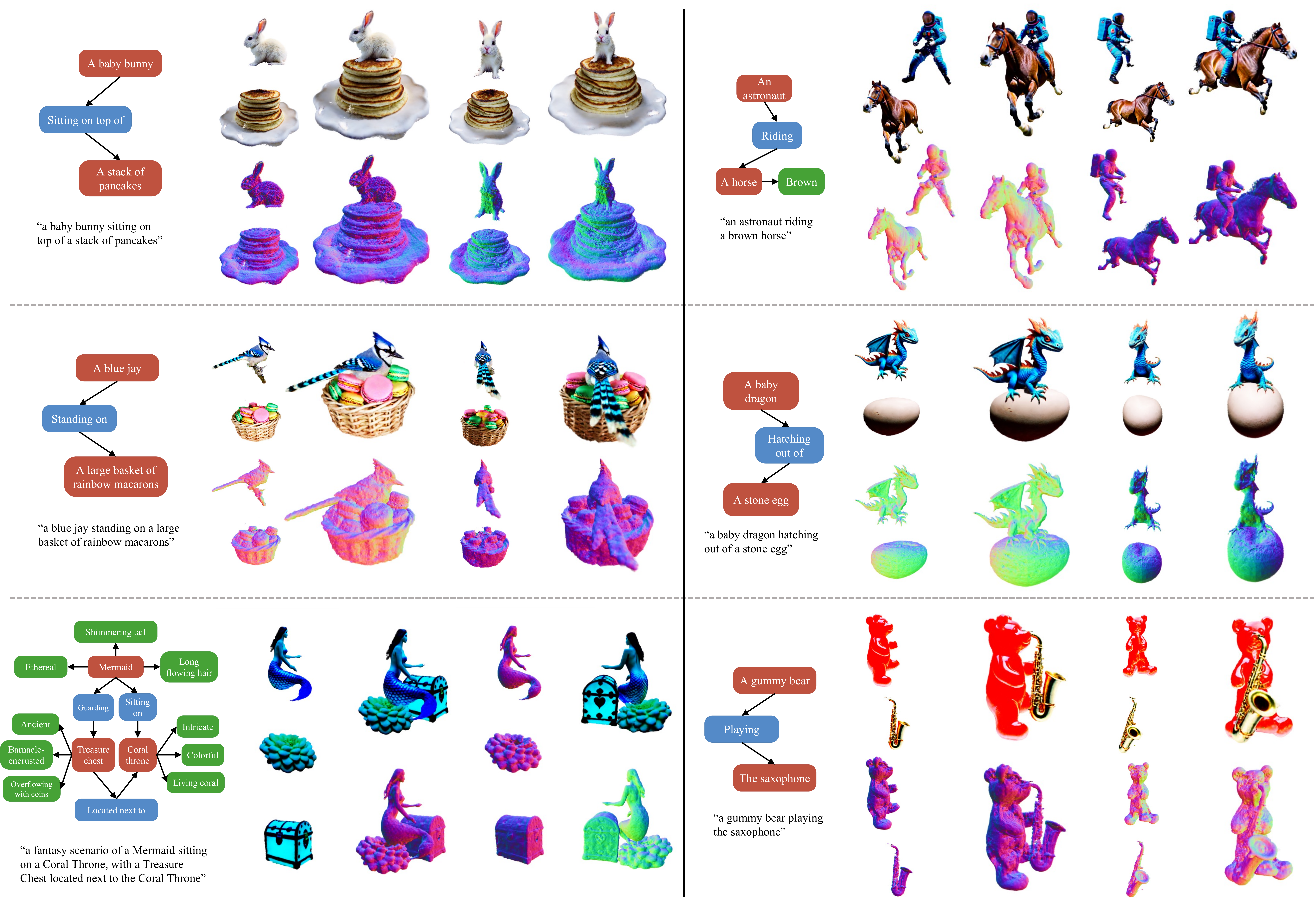}
    \vspace{-3mm}
    \caption{\footnotesize More qualitative compositional examples from GraphDreamer.}
    \label{fig:more_examples}
\vspace{-2mm}
\end{figure*}

%% file: sec/6_conclusion.tex
\section{Concluding Remarks}
This paper proposes GraphDreamer, which generates compositional 3D scenes from scene graphs or text (by leveraging GPT4-V to produce a scene graph from the text). GraphDreamer first decomposes the scene graph into global, node-wise, and edge-wise text descriptions and then optimizes the SDF-based objects with the SDS loss from their corresponding text descriptions. We conducted extensive experiments to show that GraphDreamer is able to prevent attribute confusion and guidance collapse, generating disentangled objects.

%% file: sec/7_acknowledgments.tex
\section*{Acknowledgment}
The authors extend their thanks to Zehao Yu and Stefano Esposito for their invaluable feedback on the initial draft. Our thanks also go to Yao Feng, Zhen Liu, Zeju Qiu, Yandong Wen, and Yuliang Xiu for proofreading and for their insightful suggestions on the final draft which enhanced the quality of this paper. In addition, we appreciate the assistance of those who participated in our user study. Weiyang Liu and Bernhard Sch\"olkopf was supported by the German Federal Ministry of Education and Research (BMBF): T\"ubingen AI Center, FKZ: 01IS18039B, and by the Machine Learning Cluster of Excellence, the German Research Foundation (DFG): SFB 1233, Robust Vision: Inference Principles and Neural Mechanisms, TP XX, project number: 276693517. Andreas Geiger and Anpei Chen were supported by the ERC Starting Grant LEGO-3D (850533) and the DFG EXC number 2064/1 - project number 390727645.

%% file: sec/X_suppl.tex
\clearpage
\newpage
\onecolumn
\appendix

\addcontentsline{toc}{section}{Appendix} 
\renewcommand \thepart{} 
\renewcommand \partname{}
\part{\Large{\centerline{Appendix}}}
\parttoc

\newpage

\section{Experimental Settings}

\subsection{Additional Implementation Details}

\paragraph{Point-wise identity vector $\boldsymbol{\lambda}(\bp)$.}
As defined in Eq.~(\ref{eq:soft_identity}), identity vector $\boldsymbol{\lambda}(\bp)$ at a given position $\bp$ is defined as an one-hot vector for a better disentanglement in the forward rendering process. 
To combine $\boldsymbol{\lambda}(\bp)$ into differentiate training, we customize the gradient of $\boldsymbol{\lambda}(\bp)$ in back-propagation by combining with a Softmax operation: 

\begin{equation}
\label{eq:sup_soft_identity}
    \boldsymbol{\lambda}^{+}(\bp) = \boldsymbol{\lambda}(\bp) + \Bigl\{\bs(\bp) - \bf{sg}\bigl[\bs(\bp)\bigr] \Bigr\} ~~~~~~~~ \bs(\bp) = \text{Softmax} \bigl( -\bu(\bp) \bigr)   
\end{equation}
where second term $\bs(\bp) - \bf{sg}[\bs(\bp)]$ is zero in value, with $\bf{sg}(\cdot)$ standing for the stop-gradient (\eg, \texttt{.detach()} in PyTorch) operation, and thus contributes only the gradient for updating GraphDreamer.

\paragraph{Penetration Constraint $\mathcal{L}_{penet}$.}
In Eq.~(\ref{eq:penet_loss}), we introduced a rather intuitive definition for the penetration constraint, which is, however, not continuous over object identities $i \in \{1, \cdots, M\}$. 
To further refine this constraint, we implement $\mathcal{L}_{penet}$ as:
\begin{equation}
\label{eq:sup_loss_penet} 
    \mathcal{L}_{penet} = \frac{1}{(M-1)N} \sum_{j\neq i,\bp} \Bigl\{\texttt{ReLU} \bigl[\bd(\bp) - \bu(\bp) \bigr]\Bigr\}^2  ~~~~~~~~ \bd(\bp) = \texttt{ReLU}\bigl[ \boldsymbol{\lambda}^{+}(\bp) \cdot -\bu(\bp)\bigr]
\end{equation}
where $N$ is the size of sampled positions on the current ray, $\boldsymbol{\lambda}^{+}(\bp)$ is the differentiable identity vector defined in Eq.~(\ref{eq:sup_soft_identity}), and term $\bd(\bp)$ is an one-hot vector. 
If $\bp$ is inside or on the surface of object $o^{(i)}$, and $\bu^{(i)}(\bp)$ gets the minimum among other object SDFs, the only non-negative element of $\bd(\bp)$ equals the absolute value of $\bu^{(i)}(\bp)$. 
Then, $\mathcal{L}_{penet}$ prevents all other SDF values 
($j\neq i$) from being negative. 
Or else, if $\bp$ is outside of all objects, $\mathcal{L}_{penet}$ has not impact on $\bu(\bp)$. 
An ablation study for this constraint can be found in Section~\ref{sec:sup_abla_penet}. 

\subsection{Hyperparameter Settings}

\vspace{1mm}
\noindent\textbf{Training loss coefficients}. 
We set coefficients $\{\beta_1, \beta_2,\beta_3\}$ defined in Eq.~(\ref{eq:total_loss}) based on the magnitude of each loss term, as $\beta_1 = 1$ for $\mathcal{L}_{SDS}$, $\beta_2 = 100$ for $\mathcal{L}_{penet}$, and $\beta_3 = 10$ for $\mathcal{L}_{eknl}$. 

\vspace{1mm}
\noindent\textbf{Classifier free guidance}. 
Classifier-free guidance~\cite{ho2021classifierfree} weight (CFG-w) is a hyperparameter that trades off image quality and diversity for guiding the 2D view. 
We schedule CFG-w for the training of GraphDreamer roughly based on the number of objects $M$ as below, considering that the larger the number $M$, the more difficult it is for mode seeking.

\input{00tables/sup_cfg}

\newpage
\section{Additional Experiments and Results}
\label{sec:sup_abla_penet}

\subsection{Ablation: Penetration Constraint}

As introduced in Section~\ref{sec_training} and defined practically in Eq.~(\ref{eq:sup_loss_penet}), the purpose of using the penetration constraint $\mathcal{L}_{penet}$ is to prevent unexpected penetrations between the implicit surfaces of objects, represented by SDF $u^{(i)}(\bp)$, in a multi-object scene. To verify the necessity of this constraint, we ablate $\mathcal{L}_{penet}$ in training GraphDreamer and report the quantitative and qualitative results of this ablated configuration, denoted as GraphDreamer w/o $\mathcal{L}_{penet}$, in Table~\ref{tab:supp_abla_penet} and Figure~\ref{fig:supp_abla_penet}. 
\input{figures/supp_ablation_penet}

As shown in Figure 6, objects generated without $\mathcal{L}_{penet}$ are more likely to inter-penetrate; as a result, it is much harder to identify clean segmentation boundaries between objects.
The CLIP scores shown in Table~\ref{tab:supp_abla_penet} also suggest a degradation in the performance of this ablated configuration in modeling individual objects, as the mean similarities of both object image $C^{(i)}$ with its own prompt $y^{(i)}$ (w. Self Prompt) and scene image $C^g$ with global prompt $y^g$ (Global) drop down significantly, while the mean similarity of $C^{(i)}$ with other object prompts (w. Other Prompts) increases slightly. 

\input{00tables/sup_ablation}

\input{figures/supp_more_baselines}

\subsection{User Study}
\input{00tables/sup_user_study}
To further evaluate the performance of GraphDreamer in generating guidance-compliant multi-object scenes, we conduct a survey of the results generated by the baseline approaches and GraphDreamer over $30$ multi-object text prompts. We invite $31$ raters for this study. All raters are asked to evaluate all examples and, for each prompt, to select one of the three results that semantically best fits the given prompt (if all/none of the results fit well, select the one that has the highest fidelity). The answers are summarized in Table~\ref{tab:supp_user_study}. In general, more raters ($62.26\%$) selected the results produced by GraphDreamer over the other two baseline approaches, considering our results to be more consistent with the given prompts.

\subsection{Extended Comparison to Other Text-to-3D Methods}

To provide a more comprehensive comparison on the performance of recent TT3D methods in generating multi-object scenes, we extend the quantitative and qualitative comparison reported in Table~\ref{tab:baseline_compare} and Figure~\ref{fig:baseline_compare} respectively with two more baseline approaches, DreamFusion~\cite{poole2022dreamfusion} and Fantasia3D~\cite{chen2023fantasia3d} and one more evaluation criteria, CLIP R1-Precision as reported in Table~\ref{tab:sup_more_baselines}. 
Both baseline results are obtained after $20$K training steps. For DreamFusion, we adopt the same two-stage training protocol as GraphDreamer and Magic3D, with details provided in Section~\ref{sec:experiments}. As shown in Figure~\ref{fig:supp_baseline_compare}, problems related to guidance collapse occur in both methods, while Fantasia3D failed in some cases. Note the performance of the Fantasia3D methodology may vary depending on how the shape of the SDF surface is initialized, whereas here we have only performed the default sphere initialization (with a radius of $0.5$). 
We have also conducted generation experiments with ProlificDreamer~\cite{wang2023prolificdreamer}, which adopts a three-stage optimization strategy, and yet it still failed to generated any content in these multi-object cases after $20$K steps of training in the first stage (both setting CFG-w to 100 or 200), and the results are thus not included. 

\input{00tables/sup_more_baselines}

Videos of Figure~\ref{fig:baseline_compare} and Figure~\ref{fig:more_examples} can be found in our project page.

\newpage
\section{Inverse Semantics: A Paradigm of Semantic Reconstruction with GPT4V-guided GraphDreamer}\label{app:is}

In this section, we envision a new paradigm, called \emph{inverse semantics}, which first reconstructs a scene graph from an input image and then produces a compositional 3D scene based on this scene graph. We call it inverse semantics, because it resembles the idea of \emph{inverse graphics} in a high-level sense. 
Inverse semantics differs from inverse graphics in the aspect of reconstruction that is emphasized; it focuses on interpreting semantic meaning rather than reconstructing visual details. The comparison between inverse graphics and inverse semantics is given as follows.

\input{figures/supp_inverse_semantics}

Specifically, we can implement the inverse semantics paradigm with a GPT4V-guided GraphDreamer. We first use GPT4V to obtain the scene graph from an input image, and then apply GraphDreamer to generate a compositional 3D scene based on this scene graph.
Enhanced by GPT4V's powerful image understanding capabilities, we can obtain a detailed scene graph from the input image and generate a 3D scene from the graph that semantically inverses the given image. An qualitative example of our inverse semantics paradigm is provided in Figure~\ref{fig:supp_example_inverse_3n}. 
\input{figures/supp_inverse_example_3n}

Moreover, it is also possible to extract rough estimates of the center coordinates of the objects in the graph from the image with GPT4V. The coordinate estimates are in 2D (the up-right plane) only, since the image is in 2D. For the third coordinate (on the front axis), GPT4V can offer the relative order by layering objects from the viewer's perspective. 
Then, with the SDF representation of 3D objects, 
we can use these estimates as the sphere centers when initializing object SDFs $u^{(i)}(\bp)$ to specify more accurately the spatial relationships of the objects, and thus to make the generated 3D scene better match the given image. We expect that the inverse semantics paradigm could be of general interest to the community, and our GPT4V-guided GraphDreamer serves as an important stepping stone towards this direction.

\input{figures/supp_inverse_example_5n}

\newpage
\section{Failure Cases and Limitations}

\input{figures/supp_failure_cases}

GraphDreamer still has some limitations. First, the generation quality of a single object is still largely limited by the SDS optimization. The commonly observed multi-head problems (\ie, Janus problem) may also appear in GraphDreamer. See Figure~\ref{fig:supp_failure_cases}(a) for an example. The ``piglet'' in the generated results exhibits multiple heads from different viewpoints. 
Second, the decomposition of different objects may sometimes fail. See Figure~\ref{fig:supp_failure_cases}(b) for an example. 
We observe that the disentangled ``astronaut'' still looks like a reasonable astronaut, but the disentangled violin is affected by some parts of the astronaut. 
Third, the semantic meaning of some object in the scene may be incomplete. See Figure~\ref{fig:supp_failure_cases}(c) for an example. The hero character lacks a leg in the generated result. 

We believe that part of these problems stem from the SDS loss which only offers an approximate gradient (the Jacobian of Stable Diffusion is set to identity) and the known challenges when diffusion models deal with detailed parts. We also suspect that optimizing multiple SDS losses Eq.~(\ref{eq:total_sds}) at the same time may affect disentanglement, and thus enhanced supervision on object parts and improved prompts could help.

%% file: 00tables/sup_cfg.tex
\begin{table}[h]
\footnotesize
\setlength\tabcolsep{12pt}
\renewcommand{\arraystretch}{1.2}
\centering
    \begin{tabular}[\linewidth]{c|cc|c|c}
    \textbf{Method} & \multicolumn{2}{c|}{\textbf{GraphDreamer}} & \multirow{2}{*}{\textbf{Magic3D}} & \multirow{2}{*}{\textbf{MVDream}} \\
    (Nmber of Objects) & $M = 2$ & $M \geq 3$ &  &  \\ \shline
    Coarse Stage (\textless{10K} steps) & 50 (IF) & 100 (IF) & \multicolumn{1}{c|}{100 (IF)} & \multicolumn{1}{c}{\multirow{2}{*}{50 (MVSD)}} \\
    Fine Stage (10K$\sim$20K steps) & 50 (SD) & 50 (SD) & \multicolumn{1}{c|}{50 (SD)} & \multicolumn{1}{c}{}
    \end{tabular}
\caption{\footnotesize CFG-w settings for GraphDreamer and baseline approaches. IF stands for DeepFloyd-IF model, SD for Stable diffusion, and MVSD for MVDream Stable diffusion, as detailed in Section~\ref{sec:experiments} (para. Baseline approaches).}
\label{tab:sup_cfg_w}
\end{table}

%% file: figures/supp_ablation_penet.tex
\begin{figure*}[h]
    \centering
    \includegraphics[width=\linewidth]{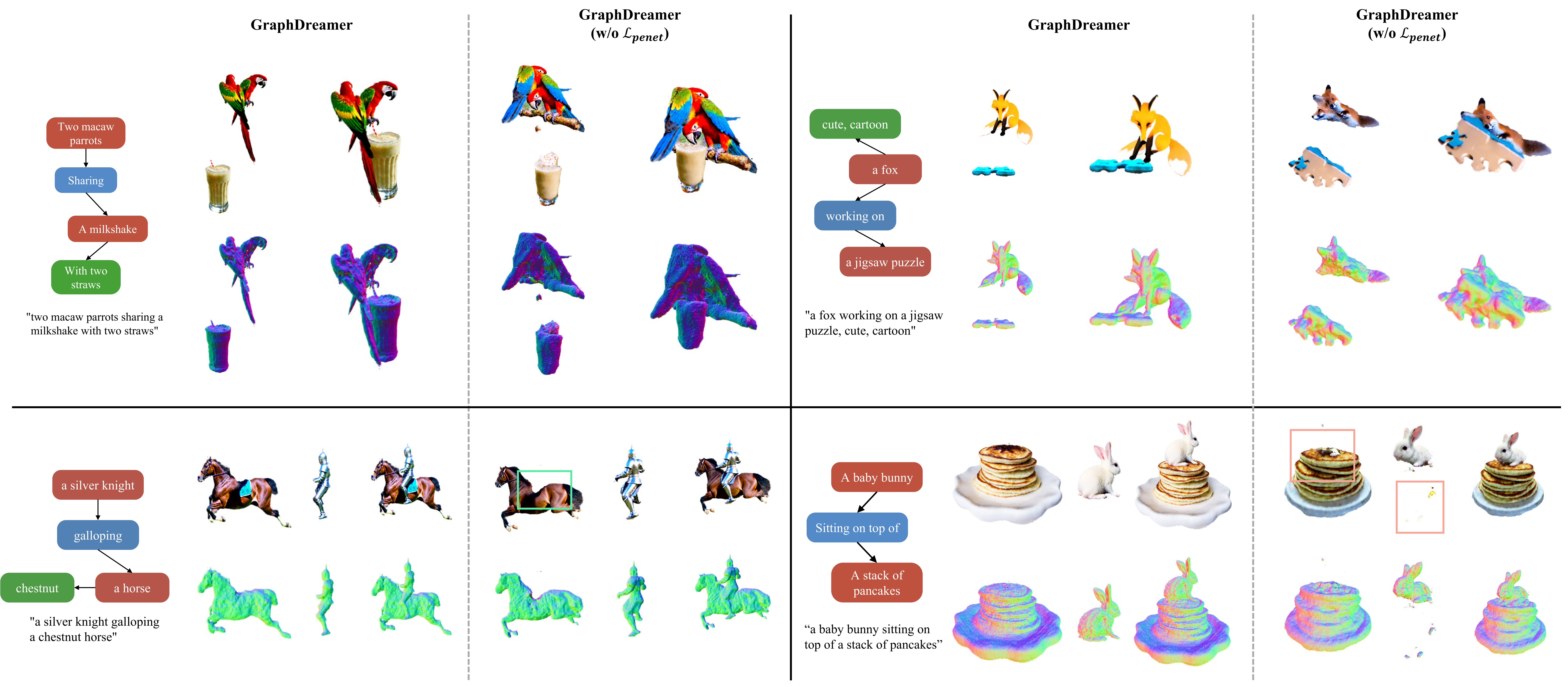}
    \caption{\footnotesize Ablation study: GraphDreamer (w/o $\mathcal{L}_{penet}$) stands for the configuration that trains GraphDreamer without penetration constraint $\mathcal{L}_{penet}(\bp)$. }
    \label{fig:supp_abla_penet}
\end{figure*}

%% file: 00tables/sup_ablation.tex
\begin{table}[h]
\setlength\tabcolsep{12pt}
\footnotesize
\renewcommand{\arraystretch}{1.2}
\centering
    \begin{tabular}[\linewidth]{c|cccc|cc}
        \multirow{2}{*}{CLIP Score} & \multicolumn{2}{c}{w. Self Prompt} & \multicolumn{2}{c|}{w. Other Prompts} & \multicolumn{2}{c}{Global} \\
         & mean ($\uparrow$) & std. & mean ($\downarrow$) & std. & mean ($\uparrow$) & std. \\ \shline
        GraphDreamer (w/o $\mathcal{L}_{penet}$) & 0.2665 & 0.0091 & 0.2070 & 0.0087 & 0.3064 & 0.0210\\
        GraphDreamer & \bf0.3077 & 0.0121 & \bf0.2006 & 0.0085 & \bf0.3326 & 0.0252
    \end{tabular}
\caption{\footnotesize Ablation study of the penetration constraint (with or without $\mathcal{L}_{penet}$).}
\label{tab:supp_abla_penet}
\end{table}

%% file: figures/supp_more_baselines.tex
\begin{figure}[t]
    \centering
    \includegraphics[width=.93\linewidth]{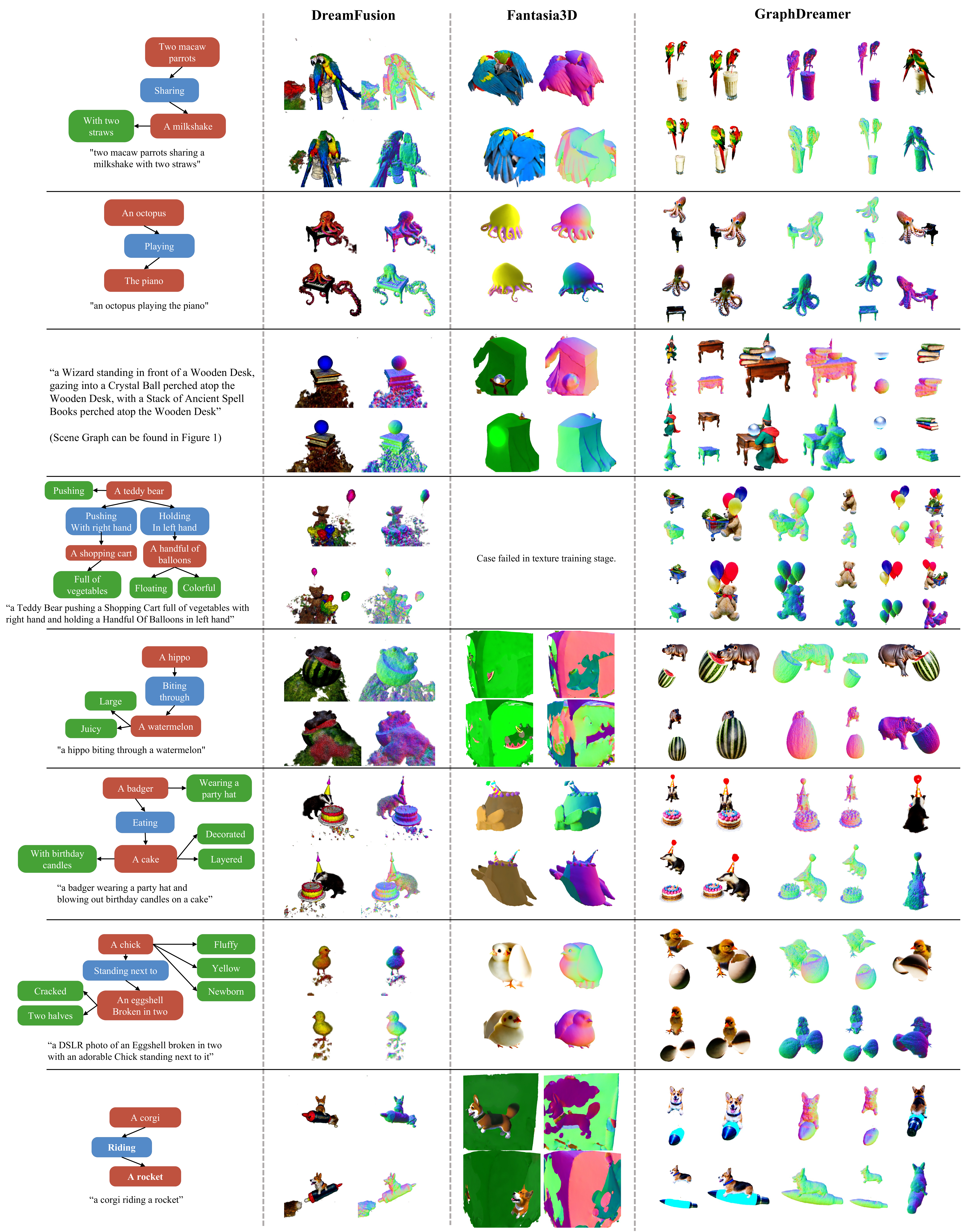}
\caption{\footnotesize Extended qualitative comparison with more baseline approaches. Note that Fantasia3D relies on the SDF initialization, which may achieve better performance if fine-grind initial shapes are provided. Here, we report results using default sphere initialization. 
}
\vspace{-5mm}
\label{fig:supp_baseline_compare}
\end{figure}

%% file: 00tables/sup_user_study.tex
\begin{table}[h]
\setlength\tabcolsep{12pt}
\footnotesize
\centering
\renewcommand{\arraystretch}{1.2}
\begin{tabular}{c|ccc}
Methods & MVDream & Magic3D & \cellcolor{Gray} GraphDreamer \\ \hline
Selected (\%) & 23.12 & 14.62 & \cellcolor{Gray} \textbf{62.26}
\end{tabular}
\caption{\footnotesize User study: selecting one from three generated results that best aligns with given text prompts. The results are collected from $31$ raters and summarized over $30$ multi-object prompts. All raters are asked to evaluate all prompt examples. }
\label{tab:supp_user_study}
\end{table}

%% file: 00tables/sup_more_baselines.tex
\begin{table}[h]
    \setlength\tabcolsep{12pt}
    \renewcommand{\arraystretch}{1.2}
    \centering
    \footnotesize
        \begin{tabular}{l|c|c|c|c|c}
         Metric & Magic3D & MVDream & DreamFusion & Fantasia3D & \textbf{GraphDreamer} \\ \shline
        CLIP R1-Precision & 87.5\% & 89.2\% & 80.0\% & 71.7\% & \cellcolor{Gray} \textbf{94.2\%} \\ 
            CLIP-B/32 Score & 0.3267 & 0.3102 & 0.2743 & 0.2243 & \cellcolor{Gray} \textbf{0.3326}
        \end{tabular}
    \caption{\scriptsize CLIP-B/32 Score \& R1-Precision across all $30$ multi-object scenes. The reported precision is averaged over 4 orthogonal views for each object. 
    }
    \label{tab:sup_more_baselines}
\end{table}

%% file: figures/supp_inverse_semantics.tex
\begin{figure}[h]
    \centering
    \includegraphics[width=.76\linewidth]{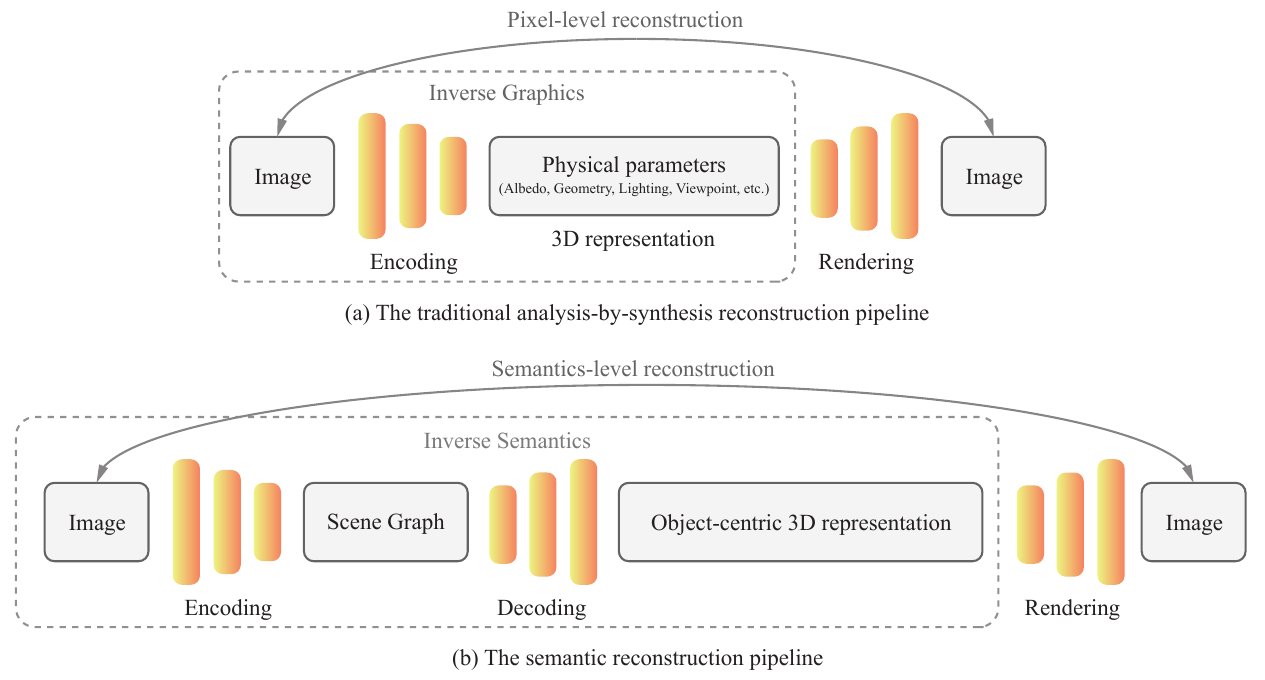}
    \vspace{-1.5mm}
    \caption{\footnotesize A paradigm comparison between inverse graphics and inverse semantics}
    \label{fig:supp_inverse_semantics}
\end{figure}

%% file: figures/supp_inverse_example_3n.tex
\begin{figure}[h]
    \centering
    \includegraphics[width=\linewidth]{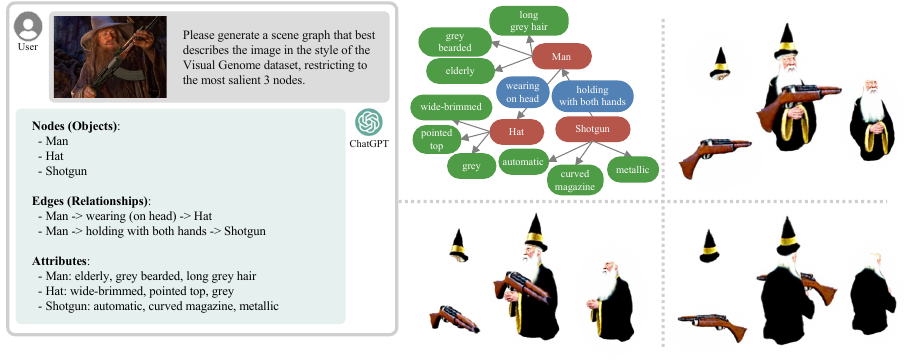}
    \vspace{-5mm}
    \caption{\footnotesize A qualitative example of the inverse semantics paradigm. We generate a scene graph directly from an input image with GPT4V and restrict the nodes present to the most salient ones. GPT4V identifies the objects and provides with proper attributes and edges for each object. This makes it possible to inverse modeling the semantics of a given scene image and extends the potential applications of GraphDreamer. }
    \label{fig:supp_example_inverse_3n}
\end{figure}

%% file: figures/supp_inverse_example_5n.tex
\begin{figure*}[h]
    \centering
    \includegraphics[width=\linewidth]{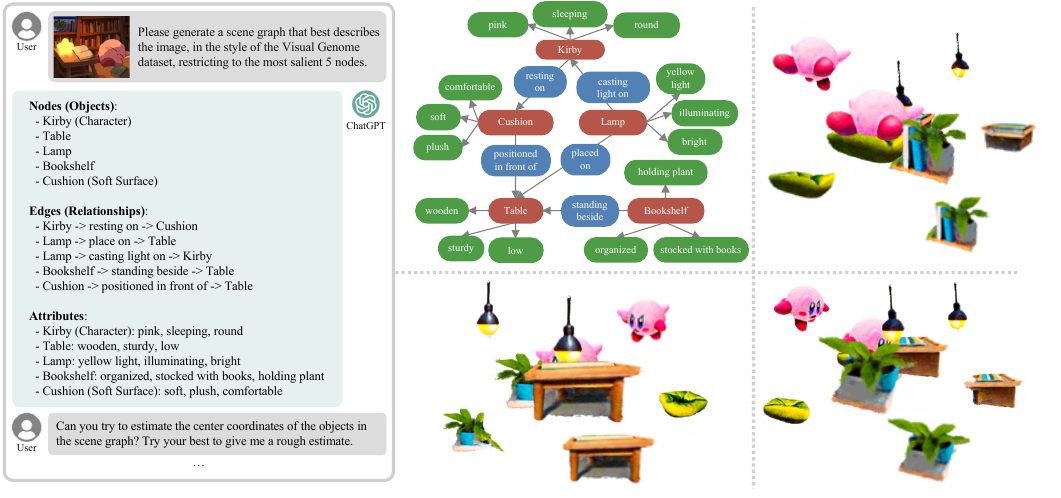}
    \caption{\footnotesize 
    An example of more complex semantics represented with more nodes. ChatGPT structures the semantics in this complex image into a scene graph using the most salient node. To inverse such semantics and generated the scene in 3D, one may further ask GPT to provide with center coordinates for each node, which can make the semantic inverse more precise in terms of spatial relations. 
    }
    \label{fig:supp_example_inverse_5n}
\end{figure*}

%% file: figures/supp_failure_cases.tex
\begin{figure*}[h]
    \centering
    \includegraphics[width=\linewidth]{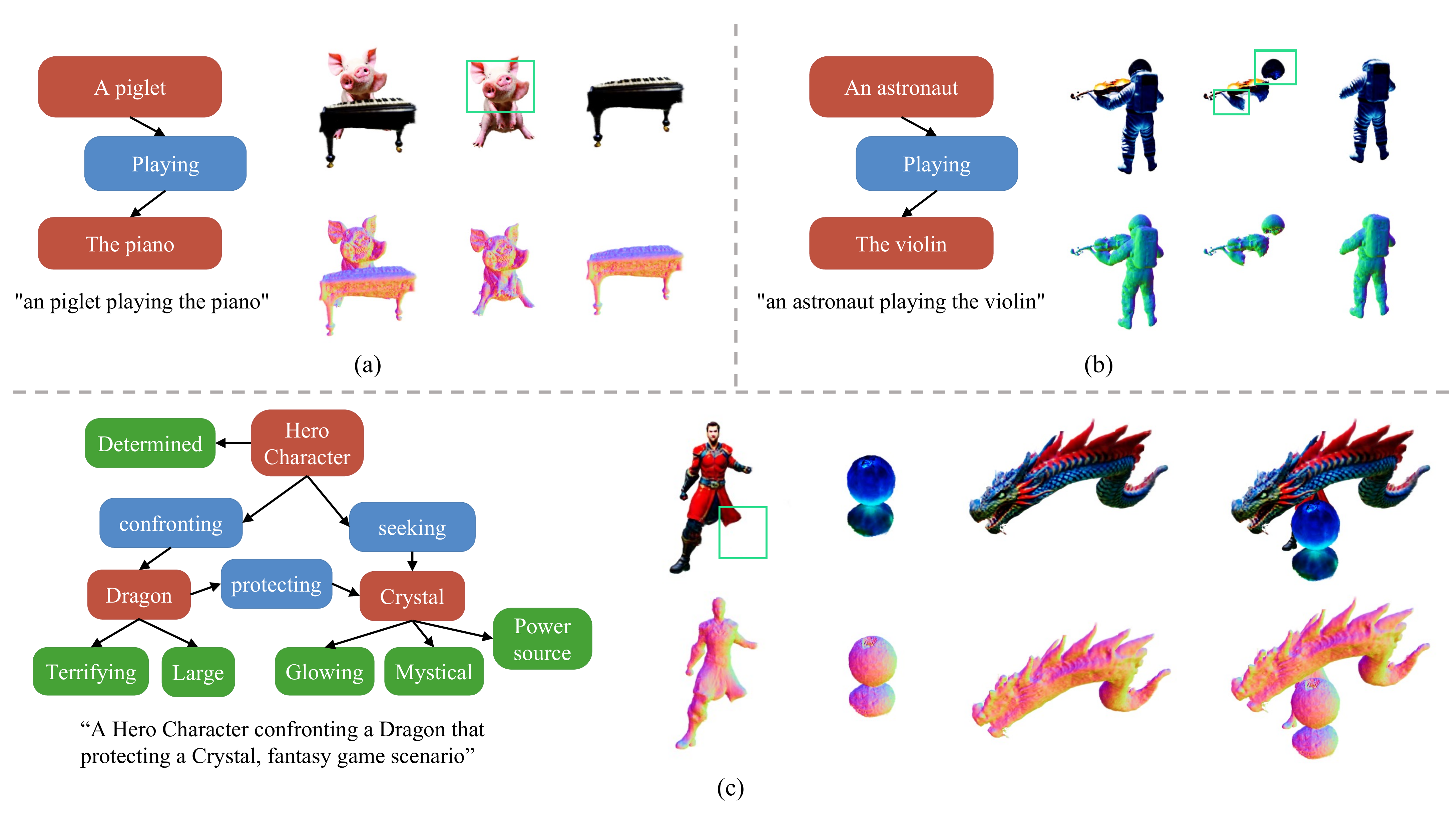}
    \caption{\footnotesize Examples of some failure cases. (a) In some cases, results of GraphDreamer may come with the \textit{Janus} problem. (b) GraphDreamer failed to separate the violin from the astronaut's left hand, which is closely held by that hand. (c) One leg of the Hero is missing. 
    Part of these problems may stem from the SDS loss which only offers an approximate gradient (the Jacobian of Stable Diffusion is set to identity) and the known challenges when diffusion models deal with detailed parts. Optimizing multiple SDS losses Eq.~(\ref{eq:total_sds}) at the same time can also affect disentanglement. Enhanced supervision on object parts and improved prompts could help. 
    }
    \label{fig:supp_failure_cases}
\end{figure*}